\documentclass[11pt,a4paper]{article}
\usepackage[hyperref]{acl2017}
\usepackage{times}
\usepackage{latexsym}
\usepackage{multirow}
\usepackage{url}

\usepackage[utf8]{inputenc}
\usepackage{amsmath}
\usepackage{amssymb}
\usepackage{wasysym}
\usepackage{color}
\usepackage{booktabs}
\usepackage{microtype}
\usepackage{tikz-dependency}
\usepackage{hhline}
\usepackage[colorinlistoftodos]{todonotes}

\usepackage{graphicx}
\graphicspath{{images/}}

\newcommand{\eat}[1]{\ignorespaces}

\usepackage{color}

\aclfinalcopy % Uncomment this line for the final submission
\def\aclpaperid{252} %  Enter the acl Paper ID here

\title{A Nested Attention Neural Hybrid Model for Grammatical Error Correction}

  \author{Jianshu Ji{$^\dagger$}, Qinlong Wang{$^\dagger$}, Kristina Toutanova{$^\ddagger$}\thanks{\ \ This work was conducted while the third author worked at Microsoft Research.}, \\ {\bf Yongen Gong{$^\dagger$}, Steven Truong{$^\dagger$}, {\bf Jianfeng Gao}{$^\mathsection$}}\\
   {{$^\dagger$}Microsoft AI \& Research  {$^\ddagger$}Google Research  {$^\mathsection$}Microsoft Research, Redmond}\\
  {{$^\dagger$}{$^\mathsection$}\tt{\{jianshuj,qinlwang,yongeg,stevetr,jfgao\}@microsoft.com}} \\{$^\ddagger$}\tt kristout@google.com}

\date{}

\begin{document}
\maketitle
\begin{abstract}

Grammatical error correction (GEC) systems strive to correct both global errors in word order and usage, and local errors in spelling and inflection. Further developing upon recent work on neural machine translation, we propose a new hybrid neural model with nested attention layers for GEC. Experiments show that the new model can effectively correct errors of both types by  incorporating word and character-level information,
and that the model significantly outperforms previous neural models for GEC as measured on the standard CoNLL-14 benchmark dataset. Further analysis also shows that the superiority of the proposed model can be largely attributed to the use of the nested attention mechanism, which has proven particularly effective in correcting local errors that involve small edits in orthography.

\end{abstract}

\section{Introduction}

One of the most successful approaches to grammatical error correction (GEC) is to cast the problem as (monolingual) machine translation (MT), where we translate from possibly ungrammatical English sentences to corrected ones \cite{brockett2006correcting,a-large-scale-ranker-based-system-for-search-query-spelling-correction,junczys2016phrase}. Such systems, which are based on phrase-based MT models that are typically trained on large sets of sentence-correction pairs, can correct global errors such as word order and usage and local errors in spelling and inflection. The approach has proven superior to systems based on local classifiers that can only fix focused errors in prepositions, determiners, or inflected forms \cite{rozovskaya-roth:2016:P16-1}.

Recently, neural machine translation (NMT) systems have achieved substantial improvements in translation quality over phrase-based MT systems \cite{sutskever2014sequence,bahdanau2014neural}. Thus, there is growing interest in applying neural systems to GEC \cite{yuan-briscoe:2016:N16-1,DBLP:journals/corr/XieAAJN16}.
In this paper, we significantly extend previous work, and explore new neural models to meet the unique challenges of GEC.

The core component of most NMT systems is a sequence-to-sequence (S2S) model which encodes a sequence of source words into a vector and then generates a sequence of target words from the vector. Unlike the phrase-based MT models, the S2S model can capture long-distance, or even global, word dependencies, which are crucial to correcting global grammatical errors and helping users achieve native speaker fluency \cite{sakaguchi2016reassessing}. Thus, the S2S model is expected to perform better on GEC than phrase-based models. However, as we will show in this paper, to achieve the best performance on GEC, we still need to extend the standard S2S model to address several task-specific challenges, which we will describe below.

First, a GEC model needs to deal with an extremely large vocabulary that consists of a large number of words and their (mis)spelling variations.

Second, the GEC model needs to capture structure  at different levels of granularity in order to correct errors of different types. For example, while correcting spelling and local grammar errors requires only word-level or sub-word level information, e.g., \emph{violets $\rightarrow$ violates} (spelling)  or \emph{violate $\rightarrow$ violates} (verb form), correcting errors in word order or usage requires global semantic relationships among phrases and words.

Standard approaches in neural machine translation, also applied to grammatical error correction by \newcite{yuan-briscoe:2016:N16-1}, address the large vocabulary problem by restricting the vocabulary to a limited number of high-frequency words and resorting to standard word translation dictionaries to provide translations for the words that are out of the vocabulary (OOV).
However, this approach often fails to take into account the OOVs in context for making correction decisions, and does not generalize well to correcting words that are unseen in the parallel training data.
An alternative approach, proposed by \newcite{DBLP:journals/corr/XieAAJN16}, applies a character-level sequence to sequence neural model. Although the model eliminates the OOV issue, it cannot effectively leverage word-level information for GEC, even if it is used together with a separate word-based language model.

\begin{figure}
    \centering
    \includegraphics[width=7.7cm]{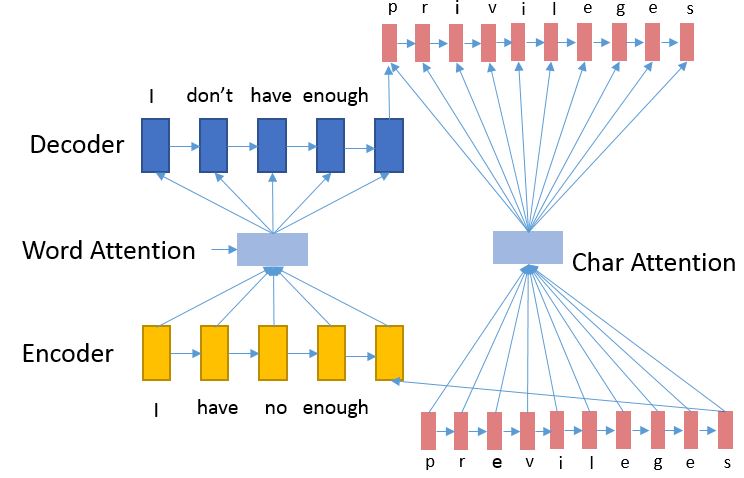}
    \caption{Architecture of Nested Attention Hybrid Model}
    \label{fig:nested model}
\end{figure}

Our solution to the challenges mentioned above is a novel, hybrid neural model with nested attention layers that infuse both word-level and character-level information. The architecture of the model is illustrated in Figure \ref{fig:nested model}.
The word-level information is used for correcting global grammar and fluency errors while the character-level information is used for correcting local errors in spelling or inflected forms.
Contextual information is crucial for GEC. Using the proposed model, by combining embedding vectors and attention at both word and character levels, we model all contextual words, including OOVs, in a unified context vector representation. In particular, as we will discuss in Section 5, the character-level attention layer captures most useful information for correcting local errors that involve small edits in orthography.

Our model differs substantially from the word-level S2S model of \newcite{yuan-briscoe:2016:N16-1} and the character-level S2S model of \newcite{DBLP:journals/corr/XieAAJN16} in the way we infuse information at both the word level and the character level. We extend the word-character hybrid model of  \newcite{luong-manning:2016:P16-1}, which was originally developed for machine translation, by introducing a character attention layer. This allows the model to learn substitution patterns at both the character level and the word level in an end-to-end fashion, using sentence-correction pairs.

%[highlight the results]
We validate the effectiveness of our model on the CoNLL-14 benchmark dataset \cite{ng2014conll}. Results show that the proposed model outperforms all previous neural models for GEC, including the hybrid model of \newcite{luong-manning:2016:P16-1}, which we apply to GEC for the first time. When integrated with a large word-based n-gram language model, our GEC system achieves an $F_{0.5}$ of  45.15 on CoNLL-14, substantially exceeding the previously reported top performance of 40.56 achieved by using a neural model and an external language model \cite{DBLP:journals/corr/XieAAJN16}.

\section{Related Work}

A variety of classifier-based and MT-based techniques have been applied to grammatical error correction. The CoNLL-14 shared task overview paper of \newcite{ng2014conll} provides a comparative evaluation of approaches. Two notable advances  after the shared task have been in the areas of combining classifiers and phrase-based MT \cite{rozovskaya-roth:2016:P16-1} and adapting phrase-based MT to the GEC task \cite{junczys2016phrase}. The latter work has reported the highest performance to date on the task of 49.5 in F$_{0.5}$ score on the CoNLL-14 test set. This method integrates discriminative training toward the task-specific evaluation function, a rich set of features, and multiple large language models. Neural approaches to the task are less explored. We believe that the advances from \newcite{junczys2016phrase} are complementary to the ones we propose for neural MT, and could be integrated with neural models to achieve even higher performance.

Two prior works explored sequence to sequence neural models for GEC \cite{DBLP:journals/corr/XieAAJN16, yuan-briscoe:2016:N16-1}, while \newcite{DBLP:journals/corr/ChollampattTN16} integrated neural features in a phrase-based system for the task. Neural models were also applied to the related sub-task of grammatical error identification \cite{schmaltz2016sentence}. \newcite{yuan-briscoe:2016:N16-1} demonstrated the promise of neural MT for GEC but did not adapt the basic sequence-to-sequence with attention to its unique challenges, falling back to traditional word-alignment models to address vocabulary coverage with a post-processing heuristic. \newcite{DBLP:journals/corr/XieAAJN16} built a character-level sequence to sequence model, which achieves open vocabulary and character-level modeling, but has difficulty with global word-level decisions.

The primary focus of our work is integration of character and word-level reasoning in neural models for GEC, to capture global fluency errors and local errors in spelling and closely related morphological variants, while obtaining open vocabulary coverage. This is achieved with the help of character and word-level encoders and decoders with two nested levels of attention.  Our model is inspired by advances in sub-word level modeling in neural machine translation. We build mostly on the hybrid model of \newcite{luong-manning:2016:P16-1} to expand its capability to correct rare words by fine-grained character-level attention. We directly compare our model to the one of \newcite{luong-manning:2016:P16-1} on the grammar correction task. Alternative methods for MT include modeling of word pieces to achieve open vocabulary \cite{sennrich-haddow-birch:2016:P16-12}, and more recently, fully character-level modeling \cite{DBLP:journals/corr/LeeCH16}. None of these models integrate two nested levels of attention although an empirical evaluation of these approaches for GEC would also be interesting.

\section{Nested Attention Hybrid Model}

Our model is hybrid, and uses both word-level and character-level representations. It consists of a word-based sequence-to-sequence model as a backbone, and additional character-level encoder, decoder, and attention components, which focus on words that are outside the word-level model's vocabulary.

\subsection{Word-based sequence-to-sequence model as backbone}

The word-based backbone closely follows the basic neural sequence-to-sequence architecture with attention as proposed by \newcite{bahdanau2014neural} and applied to grammatical error correction by \newcite{yuan-briscoe:2016:N16-1}. For completeness, we give a sketch here. It uses recurrent neural networks to encode the input sentence and to decode the output sentence.

Given a sequence of embedding vectors, corresponding to a sequence of input words  $\mathbf{x}$:
\begin{equation}
\mathbf{x} = (x_1,\ldots,x_T),
\label{eq:encoder_embedding_word}
\end{equation}

 the encoder creates a corresponding context-specific sequence of hidden state vectors $\mathbf{e}$:$$\mathbf{e}=(h_1,\ldots,h_T)$$
The hidden state $h_t$ at time $t$ is computed as: $f_t = \text{GRU}_{\text{enc}_{f}}(f_{t-1},x_t)$ ,
$b_t = \text{GRU}_{\text{enc}_{b}}(b_{t+1},x_t),$
$h_t = [f_t;b_t]$, where $\text{GRU}_{\text{enc}_{f}}$ and $\text{GRU}_{\text{enc}_{b}}$ stand for gated recurrent unit functions as described in \newcite{cho-EtAl:2014:EMNLP2014}. We use the symbol GRU with different subscripts to represent GRU functions using different sets of parameters (for example, we used the $\text{enc}_f$ and $\text{enc}_b$ subscripts to denote the parameters of the forward and backward word-level encoder  units.)

The decoder network is also an RNN using GRU units, and defines a sequence of hidden states $\bar{d_1},\ldots,\bar{d_S}$ used to define the probability of an output sequence
$y_1,\ldots,y_S$ as follows:

The context vector $c_s$ at time step $s$ is computed as follows:
\begin{equation}
c_s = \sum_{j=1}^T \alpha_{sj}h_j
\label{eq:weighted_sum_attention_word}
\end{equation}
where:
\begin{equation}
\alpha_{sk}=\frac{u_{sk}}{\sum_{j=1}^Tu_{sj}}
\label{eq:weights_attention_word}
\end{equation}
\begin{equation}
u_{sk}={\phi}_1(d_s)^T{\phi}_2(h_k)
\label{eq:feedforward_attention_word}
\end{equation}

Here ${\phi}_1$ and ${\phi}_2$ denote  feedforward linear transformations followed by a $\tanh$ nonlinearity. The next hidden state $\bar{d_s}$ is then defined as:
$$d_s = \text{GRU}_{dec}(\bar{d_{s-1}},y_{s-1}),$$
$$\bar{d_{s}} = \text{ReLU}(W[c_s;d_s])$$

where $y_{s-1}$ is the embedding of the output token at time $s$-1. ReLU indicates rectified linear units \cite{hahnloser2000digital}.

The probability of each target word $y_s$ is computed as: $ p(y_s|y_{<s},\mathbf{x}) = \text{softmax}(g(\bar{d_{s}}))$, where $g$ is a function that maps the decoder state into a vector of size the dimensionality of the target vocabulary.

The model is trained by minimizing the cross-entropy loss, which for a given $(\mathbf{x},\mathbf{y})$ pair is:
\begin{equation}
Loss(\mathbf{x},\mathbf{y})= -\sum_{s=1}^S \log p(y_s|y_{<s},\mathbf{x})
\label{eq:loss_word}
\end{equation}

For parallel training data $\mathbb{C}$, the loss is:
$$Loss = -\sum_{(\mathbf{x},\mathbf{y})\in \mathbb{C}} \sum_{s=1}^S \log p(y_s|y_{<s},\mathbf{x})$$

\subsection{Hybrid encoder and decoder with two nested levels of attention}

The word-level backbone models a limited vocabulary of source and target words, and represents out-of-vocabulary tokens with special UNK symbols. In the standard word-level NMT approach, valuable information is lost for source OOV words and target OOV words are predicted using post-processing heuristics.

\subsubsection*{Hybrid encoder}

Our hybrid architecture overcomes the loss of source information in the word-level backbone by building up compositional representations of the source OOV words using a character-level recurrent neural network with GRU units. These representations are used in place of the special source UNK embeddings in the backbone, and contribute to the contextual encoding of all source tokens.

For example, a three word input sentence where the last term is out-of-vocabulary will be represented as the following  vector of embeddings in the word-level model: $\mathbf{x} = (x_1,x_2,x_3)$, where $x_3$ would be  the embedding for the UNK symbol.

The hybrid encoder builds up a word embedding for the third word based on its character sequence: ${x^c}_1,\ldots,{x^c}_M$. The encoder computes a sequence of hidden states $\mathbf{e_c}$ for this character sequence, by a forward character-level GRU network:
\begin{equation}
\mathbf{e_c}=({h^c}_1,\ldots,{h^c}_M),
\label{eq:encoder_hidden_char}
\end{equation}

The last state ${h^c}_M$ is used as an embedding of the unknown word. The sequence of embeddings for our example three-word sequence becomes: $\mathbf{x} = (x_1,x_2,{h^c}_M)$. We use the same dimensionality for word embedding vectors $x_i$ and composed character sequence vectors ${h^c}_M$ to ensure the two ways to define embeddings are compatible. Our hybrid source encoder architecture is similar to the one proposed by \newcite{luong-manning:2016:P16-1}.

\subsubsection*{Nested attention hybrid decoder}

In traditional word-based sequence-to-sequence models special target UNK tokens are used to represent outputs that are outside the target vocabulary. A post-processing UNK-replacement method is then used \cite{cho2015using,yuan-briscoe:2016:N16-1} to replace these special tokens with target words. The hybrid model of \cite{luong-manning:2016:P16-1} uses a jointly trained character-level decoder to generate target words corresponding to UNK tokens, and outperforms the traditional approach in the machine translation task.

However, unlike machine translation, models for grammar correction conduct ``translation'' in the same language, and often need to apply a small number of local edits to the character sequence of a source word corresponding to the target UNK word. For example, rare but correct words such as entity names need to be copied as is, and local spelling errors or errors in inflection need to be corrected. The architecture of \newcite{luong-manning:2016:P16-1} does not have direct access to a source character sequence, but only uses a single fixed-dimensionality embedding of source unknown words aggregated with additional contextual information from the source.

To address the needs of the grammatical error correction task, we propose a novel hybrid decoder with two nested levels of attention: word level and character-level. The character-level attention serves to provide the decoder with direct access to the relevant source character sequence.

More specifically, the probability of each target word is defined as follows: For words in the target vocabulary, the probability is defined by the word-level backbone. For words outside the vocabulary, the probability of each token is the probability of UNK according to the backbone, multiplied by the probability of the word's character sequence.

The probability of the target character sequence corresponding to an UNK token at position $s$ in the target is defined using a character-level decoder. As in \newcite{luong-manning:2016:P16-1}, the  ``separate path'' architecture is used to capture the relevant context and define the initial state for the character-level decoder:
$$\hat{d_s} = \text{ReLU}(\hat{W}[c_s;d_s])$$
where $\hat{W}$ are parameters different from $W$, and $\hat{d_s}$ is not used by the word-level model in predicting the subsequent tokens, but is only used to initialize the character-level decoder. %for initialization state in GRU of character decoder.

To be able to attend to the relevant source character sequence when generating the target character sequence, we use the concept of hard attention \cite{xu2015show}, but use an arg-max approximation for inference instead of sampling. A similar approach to represent discrete hidden structure in a variety of architectures is used in \newcite{gragnn}.

The source index $z_s$ corresponding to the target position $s$ is defined according to the word-level attention model:

$$z_s=\arg\max_{k \in {0\ldots T-1}} \alpha_{sk}$$

where $\alpha_{sk}$ are the intermediate outputs of the word-level attention model we described in Eq.(\ref{eq:weights_attention_word}).

The character-level decoder generates a character sequence $\mathbf{y^c}=({y^c}_1,\ldots,{y^c}_N)$, conditioned on the initial vector $\hat{d_s}$ and the source index $z_s$. The characters are generated using a hidden state vector ${d^c}_n$ at each time step, via a $\text{softmax}(gc({d^c}_n))$, where $gc$ maps the state to the target character vocabulary space.

If the source word $x_{z_s}$ is in the source vocabulary, the model is analogous to the one of \newcite{luong-manning:2016:P16-1} and does not use character-level attention: the source context is available only in aggregated form to initialize the state of the decoder. The state ${d^c}_n$ for step $n$ in the character-level decoder is defined as follows, where ${\text{GRU}^c}_{\text{dec}}$ are parameters for the gated recurrent cell of this decoder:

\[{d^c}_n = \left\{
  \begin{array}{lr}
    {\text{GRU}^c}_{\text{dec}}(\hat{d_s},{y^c}_{n-1}) & n = 0 \\
    {\text{GRU}^c}_{\text{dec}}({d^c}_{n-1},{y^c}_{n-1}) &  n > 0
  \end{array}
\right.
\]

In contrast, if the corresponding token in the source $x_{z_s}$ is also an out-of-vocabulary word, we define a second nested level of character attention and use it in the character-level decoder.  The character-level attention focuses on individual characters from the source word $x_{z_s}$. If  $\mathbf{e_c}$ are the source character hidden vectors  computed as in Eq.(\ref{eq:encoder_hidden_char}), the recurrence equations for the character-level decoder with nested attention are:

$$\bar{{d^c}_n} = \text{ReLU}(W_c[{c^c}_n;{d^c}_n])$$

\[{d^c}_n = \left\{
  \begin{array}{lr}
    {\text{GRU}^c}_{\text{decNested}}(\hat{d_s},{y^c}_{n-1}) & n = 0 \\
    {\text{GRU}^c}_{\text{decNested}}(\bar{{d^c}_{n-1}},{y^c}_{n-1}) &  n > 0
  \end{array}
\right.
\]

where ${c^c}_n$ is the context vector obtained using character-level attention on the sequence $\mathbf{e^c}$ and the last state of the character-level decoder ${d^c}_n$, computed following equations \ref{eq:weighted_sum_attention_word}, \ref{eq:weights_attention_word} and \ref{eq:feedforward_attention_word}, but using a different set of parameters.

These equations show that the character-level decoder with nested attention can use both the word-level state $\hat{d_s}$, and the character-level context ${c^c}_n$ and hidden state ${d^c}_n$ to perform global and local editing operations.

Since we introduced two architectures for the character-level decoder depending on whether the source word $x_{z_s}$ is OOV, the combined loss function is defined as follows for end-to-end training:
$$Loss_{total} = Loss_w + \alpha Loss_{c1}+ \beta Loss_{c2}$$

Here $Loss_w$ is the standard word-level loss in Eq.(\ref{eq:loss_word}); character level losses $Loss_{c1}$ and $Loss_{c2}$ are losses for target OOV words corresponding to source known and unknown tokens, respectively. $\alpha$ and $\beta$ are hyper-parameters to balance the loss terms.

As seen, our proposed nested attention hybrid model uses character-level attention only when both a predicted target word and its corresponding source input word are OOV. While the model can be naturally generalized to integrate character-level attention for known words, the original hybrid model proposed by \newcite{luong-manning:2016:P16-1} does not use any character-level information for known words. Thus for a controlled evaluation of the impact of the addition of character-level attention only, in this paper we limit character-level attention to OOV words, which already use characters as a basis for building their embedding vectors. A thorough investigation of the impact of character-level information in the encoder, attention, and decoder for known words as well is an interesting topic for future research.

\subsubsection*{Decoding for word-level and hybrid models}
Beam-search is used to decode hypotheses according to the word-level backbone model. For the hybrid model architecture, word-level beam search is conducted first; for each target UNK token, character-level beam-search is used to generate a corresponding target word.

\section{Experiments}

\subsection{Dataset and Evaluation}

We use standard publicly available datasets for training and evaluation. One data source is the NUS Corpus of Learner English   (NUCLE)~\cite{dahlmeier2013building}, which is provided as a training set for the CoNLL-13 and CoNLL-14 shared tasks. From the original corpus of size about 60K parallel sentences, we randomly selected close to 5K sentence pairs for use as a validation set, and 45K parallel sentences for use in training. A second  data source is the Cambridge Learner Corpus (CLC) \cite{nicholls2003cambridge}, from which we extracted a substantially larger set of parallel sentences. Finally, we used additional training examples from the Lang-8 Corpus of Learner English v1.0 \cite{tajiri2012tense}. As Lang-8 data is crowd-sourced, we used heuristics to filter out noisy examples: we removed sentences longer than 100 words and sentence pairs where the correction was substantially shorter than the input text. Table \ref{tb:train-data} shows the number of sentence pairs from each source used for training.

We evaluate the performance of the models on the standard sets from the CoNLL-14 shared task \cite{ng2014conll}. We report final performance on the CoNLL-14 test set without alternatives, and analyze model performance on the CoNLL-13 development set \cite{dahlmeier2013building}.  We use the development and validation sets for model selection.  The sizes of all datasets in number of sentences are shown in Table \ref{tb:data}. We report performance in F$_{0.5}$-measure, as calculated by the {\texttt{m2scorer}}---the official implementation of the scoring metric in the shared task.~\footnote{\url{http://www.comp.nus.edu.sg/~nlp/sw/m2scorer.tar.gz}}
Given system outputs and gold-standard edits, {\texttt{m2scorer}} computes the $F_{0.5}$ measure of a set of system edits against a set of gold-standard edits.

\begin{table}
\centering\scalebox{.85}{
\begin{tabular}{@{}l c c c c@{}}
 & Training & Validation & Development & Test\\
%\toprule
\hline
\#Sent pairs & 2,608,679 & 4,771 & 1,381 & 1,312 \\
\hline
\end{tabular}
}
\caption{Overview of the datasets used.}
\label{tb:data}
\end{table}

\begin{table}
\centering\scalebox{.85}{
\begin{tabular}{@{}l r@{}}
 Source & \#Sent pairs \\
%\toprule
\hline
NUCLE & 45,422\\
CLC & 1,517,174\\
lang-8 & 1,046,083\\
\hline
Total & 2,608,679\\
\hline
\end{tabular}
}
\caption{Training data by source.}
\label{tb:train-data}
\end{table}

\subsection{Baseline}

We evaluate our model in comparison to the strong baseline of a word-based neural sequence-to-sequence model with attention, with post-processing for handling out-of-vocabulary words \cite{yuan-briscoe:2016:N16-1}; we refer to this model as word NMT+UNK replacement.  Like \newcite{yuan-briscoe:2016:N16-1}, we use a traditional word-alignment model (GIZA++) to derive a word-correction lexicon from the parallel training set. However, in decoding, we don't use GIZA++ to find the corresponding source word for each target OOV, but follow \newcite{cho2015using}, Section 3.3 to use the NMT system's attention weights instead. The target OOV is then replaced by the most likely correction of the source word from the word-correction lexicon, or by the source word itself if there are no available corrections.

\subsection{Training Details and Results}

The embedding size for all word and character-level encoders and decoders is set to 1000, and the hidden unit size is also 1000. To reproduce the model of \newcite{yuan-briscoe:2016:N16-1}, we selected the word vocabulary for the baseline by choosing the 30K most frequent words in the source and target respectively to form the source and target vocabularies. In preliminary experiments for the hybrid models, we found that selecting the same vocabulary of 30K words for the source and target based on combined frequency was  better (.003 in F$_{0.5}$) and use that method for vocabulary selection instead. However, there was no gain observed by using such a vocabulary selection method in the baseline. Although the source and target vocabularies in the hybrid models are the same, like in the word-level model, the embedding parameters for source and target words are not shared.

The hyper-parameters for the losses in our models are selected based on the development set and set as follows: $\alpha=\beta=0.5$. All models are trained with mini-batch size of 128 (batches are shuffled), initial learning rate of 0.0003 and a 0.95 decay ratio if the cost increases in two consecutive 100 iterations. The gradient is rescaled whenever its norm exceeds 10, and dropout is used with a probability of 0.15. Parameters are uniformly initialized in $[ -\frac{\sqrt(3)}{\sqrt{1000}} , \frac{\sqrt(3)}{\sqrt{1000}}]$.

We perform inference on the validation set every 5000 iterations to log word-level cost and character-level costs; we save parameter values for the model every 10000 iterations as well as the end of each epoch. The stopping point for training is selected based on development set $F_{0.5}$ among the top 20 parameter sets with best validation set value of the loss function.

Training of the nested attention hybrid model takes approximately five days on a Tesla k40m GPU. The basic hybrid model trains in around four days and the word-level backbone trains in approximately three days.

Table \ref{tb:pure_final} shows the performance of the baseline and our nested attention hybrid model on the development and test sets. In addition to the word-level baseline, we include the performance of a hybrid model with a single level of attention, which follows the work of \newcite{luong-manning:2016:P16-1} for machine translation, and is the first application of a hybrid word/character-level model to grammatical error correction. Based on hyper-parameter selection, the character-level component weight of the loss is $\alpha=1$ for the basic hybrid model.

As shown in Table \ref{tb:pure_final}, our implementation of the word NMT+UNK replacement baseline approaches the performance of the one reported in \newcite{yuan-briscoe:2016:N16-1} (38.77 versus 39.9). We attribute the difference to differences in the training set and the word-alignment methods used. Our reimplementation serves to provide a controlled experimental evaluation of the impact of hybrid models and nested attention on the GEC task. As seen, our nested attention hybrid model substantially improves upon the baseline, achieving a gain of close to 3 points on the test set. The hybrid word/character model with a single level of attention brings a large improvement as well, showing the importance of character-level information for this task. We delve deeper into the impact of nested attention for the hybrid model in Section 5.

\eat{
\begin{table}[t]
    \centering
     \scalebox{.9}{
    \begin{tabular}{lcc}
    \toprule
     % &  \multicolumn{2}{c}{\textbf{Drug-Gene-Mutation}} & \multicolumn{2}{c}{\textbf{Drug-mutation}} \\
    \textbf{\small Model}  & \multicolumn{2}{c}{\textbf{\small Performance}} \\ %& \textbf{Single-Sent.} & \textbf{Cross-Sent.} \\
      &  Dev & Test \\
    \midrule
        {\small Word NMT + UNK replacement} & 26.17 & 38.77 \\  % & 73.9 & 75.2 \\
        {\small Nested Attention Hybrid Model} & 28.61 & 41.53\\  % & 73.9 & 75.2 \\
    \bottomrule
    \end{tabular}
    }
    \caption{
        F$_{0.5}$ results on the CoNLL-13 and CoNLL-14 test sets of main model architectures.
        }
    \label{tb:pure_final}
\end{table}
}
\begin{table}[t]
    \centering
    \scalebox{.85}{
    \begin{tabular}{lcc}
    \toprule
     % &  \multicolumn{2}{c}{\textbf{Drug-Gene-Mutation}} & \multicolumn{2}{c}{\textbf{Drug-mutation}} \\
    \textbf{ Model}  & \multicolumn{2}{c}{\textbf{ Performance}} \\ %& \textbf{Single-Sent.} & \textbf{Cross-Sent.} \\
      &  Dev & Test \\
    \midrule
        { Word NMT + UNK replacement} & 26.17 & 38.77 \\  % & 73.9 & 75.2 \\
        { Hybrid model } & 28.49 & 40.44 \\  % & 73.9 & 75.2 \\
        { Nested Attention Hybrid Model} & \textbf{28.61} & \textbf{41.53}\\  % & 73.9 & 75.2 \\
    \bottomrule
    \end{tabular}
    }
    \caption{
        { F$_{0.5}$ results on the CoNLL-13 and CoNLL-14 test sets of main model architectures.}
        }
    \label{tb:pure_final}
\end{table}

\subsection{Integrating a Web-scale Language Model}

The value of large language models for grammatical error correction is well known, and such models have been used in classifier and MT-based systems. To establish the potential of such models in word-based neural sequence-to-sequence systems, we integrate a web-scale count-based language model. In particular, we use the modified Kneser-Ney 5-gram language model trained from  Common Crawl~\cite{buck2014n}, made available for download by \newcite{junczys2016phrase}.

Candidates generated by neural models are re-ranked using the following linear interpolation of log probabilities: $s_{y|x} = \log P_{NN}(y|x) + \lambda \log P_{LM}(y)$. Here $\lambda$ is a hyper-parameter that balances the weights of the neural network model and the language model. We tuned $\lambda$ separately for each neural model variant, by exploring values in the range $[0.0,2.0]$ with step size $0.1$, and selecting according to development set F$_{0.5}$. The selected values of $\lambda$ are: $1.6$ for word NMT + UNK replacement and $1.0$ for the nested attention model.  

Table \ref{tb:final} shows the impact of the LM when combined with the neural models implemented in this work. The table also lists the results reported by \newcite{DBLP:journals/corr/XieAAJN16}, for their character-level neural model combined with a large word-level language model. Our best results exceed the ones reported in the prior work by more than 4 points, although we should note that \newcite{DBLP:journals/corr/XieAAJN16} used a smaller parallel data set for training.

\begin{table}[t]
    \centering
    \scalebox{.85}{
    \begin{tabular}{lp{0.7cm}p{0.7cm}}
    \toprule
     % &  \multicolumn{2}{c}{\textbf{Drug-Gene-Mutation}} & \multicolumn{2}{c}{\textbf{Drug-mutation}} \\
    \textbf{\small Model}  & \multicolumn{2}{c}{\textbf{\small Performance}} \\ %& \textbf{Single-Sent.} & \textbf{Cross-Sent.} \\
      &  Dev & Test \\
    \midrule

        {\small Character-based NMT + LM \cite{DBLP:journals/corr/XieAAJN16}} & & 40.56 \\ \hline  % & 73.9 & 75.2 \\
        {\small Word NMT + UNK replacement + LM} & 31.73 & 42.82 \\  % & 73.9 & 75.2 \\
        {\small Hybrid model + LM} & 33.21 & 44.99  \\  % & 73.9 & 75.2 \\
        {\small Nested Attention Hybrid Model + LM} & \textbf{33.47}  & \textbf{45.15} \\  % & 73.9 & 75.2 \\
    \bottomrule
    \end{tabular}
    }
    \caption{
        F$_{0.5}$ results on the CoNLL-13 and CoNLL-14 test sets of main model architectures, when combined with a large language model.
        }
    \label{tb:final}
\end{table}

\eat{
\begin{table}[t]
    \centering
    \begin{tabular}{p{5.5cm}c}
    \toprule
     % &  \multicolumn{2}{c}{\textbf{Drug-Gene-Mutation}} & \multicolumn{2}{c}{\textbf{Drug-mutation}} \\
    \textbf{\small Model}  & \multicolumn{1}{c}{\textbf{\small Performance}} \\ %& \textbf{Single-Sent.} & \textbf{Cross-Sent.} \\
      &  Test \\
    \midrule

        {\small character-based NMT + LM \cite{DBLP:journals/corr/XieAAJN16}} & 40.56 \\ 
        {\small Nested Attetion Hybrid Model+LM} & 45.15 \\  % & 73.9 & 75.2 \\
        
    \bottomrule
    \end{tabular}
    \caption{
        F$_{0.5}$ results on the CoNLL-14 test sets of main model architectures, when combined with a large language model.
        }
    \label{tb:final}
\end{table}
}

\section{Analysis}

We analyze the impact of sub-word level information and the two nested levels of attention in more detail by looking at the performance of the models on different segments of the data. In particular, we analyze the performance of the models on sentences containing OOV source words versus ones without OOV words, and corrections to orthographically similar versus dissimilar word forms.

\subsection{Performance by Segment: OOV versus Non-OOV}

We present a comparative performance analysis of models on the CoNLL-13 development set. First, we divide the set into two segments: OOV and NonOOV, based on whether there is at least one OOV word in the given source input.  Table \ref{tb:OOV_NonOOV_sent}  shows that both hybrid architectures substantially outperform the word-level model in both segments of the data. The additional nested character-level attention of our hybrid model brings a sizable improvement over the basic hybrid model in the OOV segment and a small degradation in the non-OOV segment. We should note that in future work character-level attention can be added for non-OOV source words in the nested attention model, which could improve performance on this segment as well.

\begin{table}[t]
    \centering
    \scalebox{.85}{
    \begin{tabular}{p{4cm}p{1cm}p{0.7cm}p{0.7cm}}
    \toprule
     % &  \multicolumn{3}{c}{\textbf{Drug-Gene-Mutation}} & \multicolumn{3}{c}{\textbf{Drug-mutation}} \\
    \textbf{\small Model} & \textbf{\small NonOOV} & \textbf{\small OOV} & \textbf{\small Overall} \\ %\multicolumn{3}{c}{\textbf{\small Segments}} \\ %& \textbf{Single-Sent.} & \textbf{Cross-Sent.} \\
    %& \small NonOOV & \small OOV & \small Overall \\
    \midrule

        {\small Word NMT + UNK replacement} &  27.61 &  21.57 &  26.17\\  % & 73.9 & 75.2 \\ \hline
        {\small Hybrid model} &  \textbf{29.36} &  25.92 & 28.49\\ %\hline  % & 73.9 & 75.2\\ \hline
        {\small Nested Attention Hybrid Model} & 29.00 & \textbf{27.39} & \textbf{28.61}\\  % & 73.9 & 75.2\\
    \bottomrule
    \end{tabular}
    }
    \caption{
        F$_{0.5}$ results on  the CoNLL-13 set of main model architectures, on different segments of the set according to whether the input contains OOVs.
        }
    \label{tb:OOV_NonOOV_sent}
\end{table}

Table \ref{tb:case_OOV} shows an example where the nested attention hybrid model successfully corrects a misspelling resulting in an OOV word on the source, whereas the baseline word-level system simply copies the source word without fixing the error (since this particular error is not observed in the parallel training set).

\begin{table}[t]
    \centering
    \scalebox{.8}{
    \begin{tabular}{p{2.5cm}|p{4.5cm}}
    \toprule
    \multirow{4}{*} \small{source} & This greatly \textbf{violets} the rights of  people .\\ \hline
    \small{gold} & This greatly \textbf{violates} the rights of people .\\ \hline
    %& word NMT & This greatly \_ UNK the rights of people . \\ \cline{2-3}
    \small{word NMT + UNK replacement} & This greatly \textbf{violets} the rights of people .\\ %\hline \hline
    \hline
    %& \small{Hybrid Model} & This greatly \textbf{violates} the rights of people .\\\cline{2-3}
    \small{Nested Attention Hybrid Model} & This greatly \textbf{violates} the rights of people .
   \\ %\hline \hline

    \bottomrule
    \end{tabular}
    }
    \caption{
        An example sentence from the OOV segment where the nested attention hybrid model improves performance.
        }
    \label{tb:case_OOV}
\end{table}

\subsection{Impact of Nested Attention on Different Error Types}

To analyze more precisely the impact of the additional character-level attention introduced by our design, we continue to investigate the OOV segment in more detail.

The concept of \emph{edit}, which is also used by the official M2 score metric, is defined as a minimal pair of corresponding sub-strings in a source sentence and a correction. For example, in the sentence fragment pair: {``Even though there is a risk of causing \textbf{harms} to someone, people still \textbf{are prefers} to keep their pets without a leash.''} $\rightarrow$ {``Even though there is a risk of causing \textbf{harm} to someone, people still \textbf{prefer} to keep their pets without a leash.''}, the minimal edits are ``harms $\rightarrow$ harm'' and ``are prefers $\rightarrow$ prefer''. The $F_{0.5}$ score is computed using weighted precision and recall of the set of a system's edits against one or more sets of reference edits.

For our in-depth analysis, we classify edits in the OOV segment into two types: \emph{small changes} and \emph{large changes}, based on whether the source and target phrase of the edit are orthographically similar or not. More specifically, we say that the target and source phrases are orthographically similar, iff: the character edit distance is at most 2 and the source or target is at most 8 characters long, or $edit\_ratio < 0.25$, where $edit\_ratio = \frac{ {character\_edit\_distance}}{ \min(len(src),len(tar))+0.1}$, $len(*)$ denotes number of characters in $*$, and $src$ and $tgt$ denote the pairs in the edit. There are 307 gold edits in the ``small changes'' portion of the CoNLL-13 OOV segment, and 481 gold edits in the ``large changes'' portion.

Our hypothesis is that the additional character-level attention layer is particularly useful to model edits among orthographically similar words. Table \ref{tb:Orthographically_similar} contrasts the impact of character-level attention on the two portions of the data. We can see that the gains in the ``small changes'' portion are indeed quite large, indicating that the fine-grained character-level attention empowers the model to more accurately correct confusions among phrases with high character-level similarity. The impact in the ``large changes'' portion is slightly positive in precision and slightly negative in recall. Thus most of the benefit of the additional character-level attention stems from improvements in the ``small changes'' portion.

\begin{table}[t]
    \centering
    \begin{tabular}{p{3.9cm}p{0.6cm}p{0.55cm}p{0.55cm}}
    \toprule
     % &  \multicolumn{3}{c}{\textbf{Drug-Gene-Mutation}} & \multicolumn{3}{c}{\textbf{Drug-mutation}} \\
    \textbf{\small Model} &
    \multicolumn{3}{c}{\textbf{\small Performance}} \\ %& \textbf{Single-Sent.} & \textbf{Cross-Sent.} \\

      & P & R & $F_{0.5}$ \\
      \hline
    \midrule
     \multicolumn{4}{c}{\textbf{\small Small Changes Portion}} \\
        {\small Hybrid model } & 43.86 & 16.29 & 32.77 \\  % & 73.9 & 75.2\\
        {\small Nested Attention Hybrid Model} & 48.25 & 17.92 & 36.04\\  % & 73.9 & 75.2 \\
    \midrule
    \hline
    \multicolumn{4}{c}{\textbf{\small Large Changes Portion}} \\
        {\small Hybrid model } & 32.52 & 8.32 & 20.56 \\  % & 73.9 & 75.2\\
        {\small Nested Attention Hybrid Model} & 33.05 & 8.11 & 20.46\\  % & 73.9 & 75.2 \\
    \bottomrule
    %\bottomrule
    \end{tabular}
    \caption{
        Precision, Recall and $F_{0.5}$ results on CoNLL-13,on the "small changes" and ``large changes'' portions of the OOV segment.
        }
    \label{tb:Orthographically_similar}
\end{table}

Table \ref{tb:case} shows an example input which  illustrates the precision gain of the nested attention hybrid model. The input sentence has a source OOV word which is correct. The hybrid model introduces an error in this word, because it uses only a single source context vector, aggregating the character-level embedding of the source OOV word together with other source words. The additional character-level attention layer in the nested hybrid model enables the correct copying of this long source OOV word, without employing the heuristic mechanism of the word-level NMT system.

\begin{table}[t]
    \centering
    \scalebox{.85}{
    \begin{tabular}{p{2.5cm}|p{4.5cm}}
    \toprule
    \multirow{5}{*} \small{source} & Population ageing : A more and more \textbf{attention-getting} topic\\ \hline%\cline{2-3}
     \small{gold} & Population ageing : A more and more \textbf{attention-getting} topic\\ \hline%\cline{2-3}
     \small{Word NMT + UNK replacement} & Population ageing : A more and more \textbf{attention-getting} topic\\ \hline %\hline
    %\cline{2-3}
     \small{Hybrid Model} & Population ageing : A more and more \textbf{attention-teghting} topic\\ \hline
     \small{Nested Attention Hybrid Model} & Population ageing : A more and more \textbf{attention-getting} topic\\ %\hline%\hline \hline
    %\multirow{4}{*}{2} & source & Since it is \textbf{read-only} , the card can not protect itself .\\ \cline{2-3}
    %& gold & Since it is \textbf{read-only} , the card can not protect itself .\\ \cline{2-3}
    %& hybrid model & Since it is \textbf{readonly} , the card can not protect itself .\\ \cline{2-3}
    %& nested model  & Since it is \textbf{read-only} , the card can not protect itself .\\ %\cline{2-3}
    \bottomrule
    \end{tabular}
    }
    \caption{
        An example where the nested attention hybrid model outperforms the non-nested model.
        }
    \label{tb:case}
\end{table}

\section{Conclusions}
We have introduced a novel hybrid neural model with two nested levels of attention: word-level and character-level. The model addresses the unique challenges of the grammatical error correction task and achieves the best reported results on the CoNLL-14 benchmark among fully neural systems. Our nested attention hybrid model deeply combines the strengths of word and character level information in all components of an end-to-end neural model: the encoder, the attention layers, and the decoder. This enables it to correct both global word-level and local character-level errors in a unified way. The new architecture contributes substantial improvement in correction of confusions among rare or orthographically similar words compared to word-level sequence-to-sequence and non-nested hybrid models.

\section*{Acknowledgements}

We would like to thank the ACL reviewers for their insightful suggestions, Victoria Zayats for her help with reproducing the baseline word-level NMT system and Yu Shi, Daxin Jiang and Michael Zeng for the helpful discussions.

%\bibliography{bibliography-file}
%\bibliographystyle{acl_natbib}
\bibliography{acl2016}
\bibliographystyle{acl_natbib}
\end{document}